\documentclass[compsoc, conference, letterpaper, 10pt, times]{IEEEtran}
\usepackage{spconf,amsmath,graphicx}
\usepackage{float}
\usepackage{subfig}
\usepackage{graphicx}
\ifCLASSOPTIONcompsoc
  \usepackage[nocompress]{cite}
\else
  \usepackage{cite}
\fi
\ifCLASSINFOpdf
\else
\fi

\usepackage{url}

\title{Removal of Batch Effects using Generative Adversarial Networks}
\name{Uddeshya Upadhyay, Arjun Jain}
\address{Department of Computer Science and Engineering\\
Indian Institute of Technology-Bombay\\}

\begin{document}
\maketitle
\begin{abstract}
Many biological data analysis processes like Cytometry or Next Generation Sequencing (NGS) produce massive amounts of data which needs to be processed in batches for down-stream analysis. Such datasets are prone to technical variations due to difference in handling the batches possibly at different times, by different experimenters or under other different conditions. This adds variation to the batches coming from the same source sample. These variations are known as Batch Effects. It is possible that these variations and natural variations due to biology confound but such situations can be avoided by performing experiments in a carefully planned manner. Batch effects can hamper down-stream analysis and may also cause results to be inconclusive. Thus, it is essential to correct for these effects. This can be solved using a novel Generative Adversarial Networks (GANs) based framework that is proposed here, advantage of using this framework over other prior approaches is that here it is not required to choose a reproducing kernel and define its parameters. Results of the framework on a mass cytometry dataset are reported.
\end{abstract}
\section{Introduction and Related Work}
\label{sec:intro}
Batch effects are technical sources of variation that have been added to the samples during handling. They are common in many biological data analysis pipelines as many such experiments require the sample to be divided in different batches and such effects might get introduced at the time of creation of these batches due to variation in environmental conditions such as temperature, instruments or other experimenter related conditions. Although these are not the only sources of variation which may cause the batches to differ but not correcting for them will lead to different outputs from different batches from same source sample and the experiments will be rendered inconclusive.

Currently intuition and domain knowledge from the expert side is used to identify the underlying candidate parameters which might have caused such batch effects. One of the preliminary tests to check if a particular underlying variable is the cause for variation is to plot the projection of the batches on few principle components and mark them differently on the basis of different value of concerned variable, if the points separate out then clearly the given variable is responsible for the effect \cite{doi:10.1093/bioinformatics/btt480}.

\section{Related Work}
\label{sec:related}
Some of the recent deep learning based methods to solve this problem utilize residual networks to learn a near identity mapping from source to target by optimizing the Maximum Mean Discrepancy (MMD) between the transformed source and original target \cite{shaham2017removal,Dziugaite:2015:TGN:3020847.3020875,li2015generative}. MMD is one the several methods used to quantify the distance between two continuous distributions. It uses the distance measure between the means of distribution in a transformed space as a proxy to distance measure between distributions in original space. Let $P$ and $Q$ be two distributions over set $A$, let $\phi : A \xrightarrow{}H$ be a transformation from $A$ to reproducing kernel Hilbert space $H$ then MMD is defined as
\begin{align}
    \textrm{MMD(P, Q)} = \|\mathrm{E_{X_1 \sim P}(\phi(X_1))} - \mathrm{E_{X_2 \sim Q}(\phi(X_2))}\|
\end{align}
MMD depends upon the choice of reproducing kernel and hence one needs to devise a method to find the optimum kernel parameters.
Proposed solution is based on a Generative Adversarial Network (GAN) and results are shown on a mass cytometry dataset which was used for a similar study before. This method does not involve reproducing kernel, therefore one does not have to specify and discover kernel parameters explicitly.

\section{Methods}
\label{sec:methods}
\subsection{Visualizing Batch Effects}
\begin{figure*}
\minipage{0.32\textwidth}
  \includegraphics[width=\linewidth, height=\linewidth]{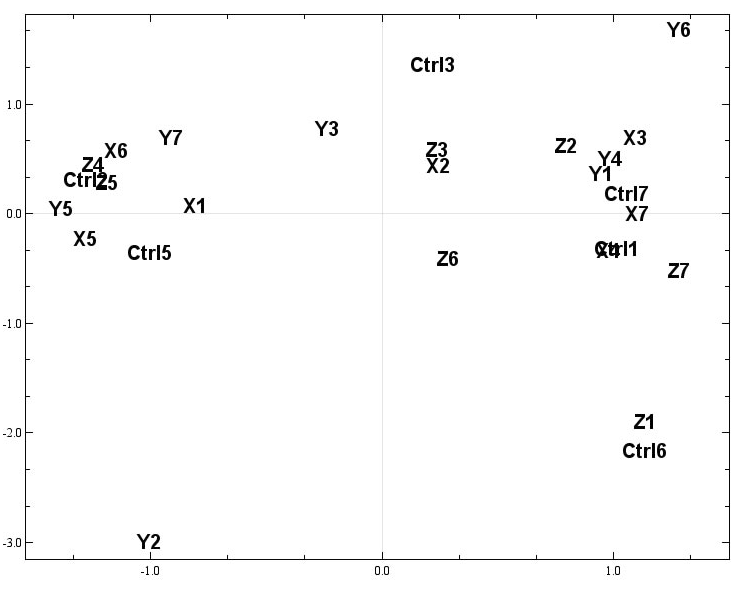}
  \centering{(a)}
\endminipage\hfill
\minipage{0.32\textwidth}
  \includegraphics[width=\linewidth, height=\linewidth]{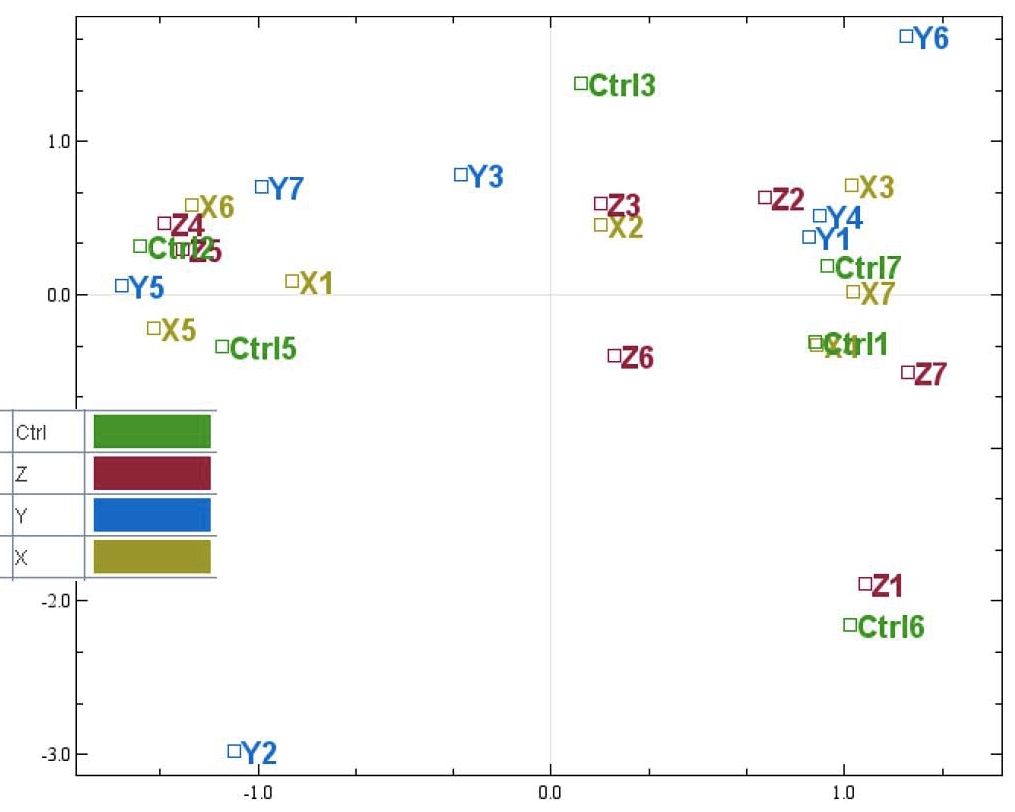}
  \centering{(b)}
\endminipage\hfill
\minipage{0.32\textwidth}%
  \includegraphics[width=\linewidth, height=\linewidth]{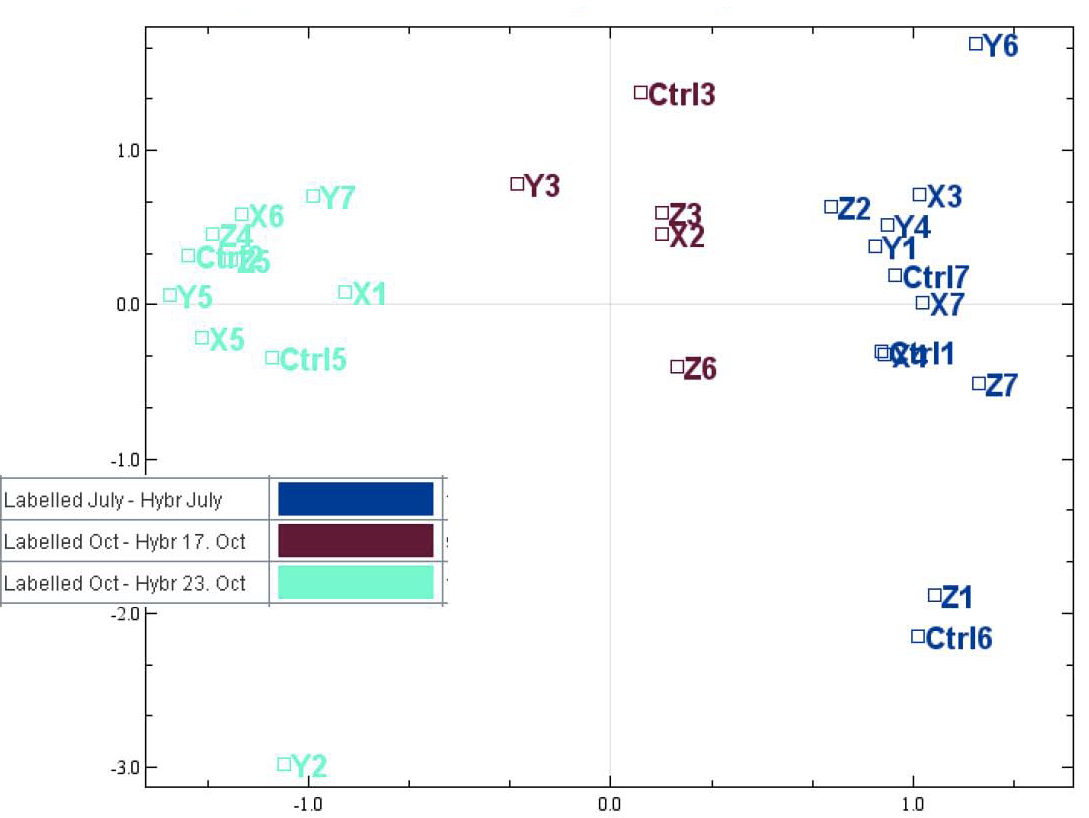}
  \centering{(c)}
\endminipage
\caption{PCA plots showing samples from a dataset consisting of four biological groups. (a) shows the projections without any markers, (b) shows projections marked according to their biological group, it is clear that there are 3 major clusters and each cluster has good mix of every biological groups. (c) shows projections marked with sampling dates and it is evident that date is the factor introducing batch effects}
\label{fig:fig1}
\end{figure*}

Figure~\ref{fig:fig1} shows an example where a dataset consisting various biological groups has been plotted in 2-dimensional space by taking the projection on first two major principle components. Figure~\ref{fig:fig1}(a) shows that while plotting the population without any markers does not reveal a discernible pattern while plotting the same points marked with colour representing their group (in this case population consists of four different groups) shows that each group is well mixed in the three different clusters distinctly visible. The underlying cause of such clustering is sampling date and this is clear if we plot the points marking each point according to sampling dates. The three different dates correspond to three different clusters\footnote{Visualization taken from \url{http://www.molmine.com/magma/global_analysis/batch_effect.html}}.
In general plotting projections along a few principle components and marking the points according to different underlying variables can help detect the cause of batch effects. However, it requires the domain knowledge and intuition of an analyst to hypothesize which variable might be causing such effects.

\subsection{Generative Models}
Generative models take training data i.e. samples from $P_{data}$ (which is unknown) and learn to represent the unknown distribution. Some models do this explicitly by estimating the unknown distribution using $P_{model}$. This can be done in many ways such as modelling $P_{data}$ by a family of known distributions and then estimating the parameters of the family by optimization techniques such as Maximum Likelihood Estimate (MLE). Other such explicit methods are Variational Autoencoders, Boltzmann Machines, etc. 
Often it suffices to not have an explicit representation of $P_{data}$ in the form of $P_{model}$ but to be able to produce samples from $P_{model}$ which approximate $P_{data}$. Generative Adversarial Networks (GANs) are one such approach where $P_{model}$ does not estimate $P_{data}$ directly, but the trained network can generate new samples from $P_{model}$ which approximates $P_{data}$ as explained in \cite{gan_tut}. The following sections provide details of how this indirect method can be useful for this problem.

\subsection{Generative Adversarial Networks}
Generative Adversarial Network (GAN) \cite{goodfellow2014generative} is a framework where two artificial neural networks compete against each other. One of them is a \textit{Generator ($G$)} with parameters $\theta_G$ and the other is a \textit{Discriminator ($D$)} with parameters $\theta_D$. Typically generator takes an input $z \sim P_{prior}$ where $P_{prior}$ is a simple prior distribution like Gaussian and outputs a value $x \sim P_{model}$ where $P_{model}$ tries to approximate $P_{data}$. Generator minimizes the cost function $C_{G}(.;\theta_{G}, \theta_{D})$ which depends upon the parameters of $G$ as well as $D$.

The Discriminator takes input $x \sim P_{data} \cup P_{model}$, i.e. inputs are sampled from both training data and the output produced by generator. Let $x_{real} \sim P_{data}$ and $x_{fake} \sim P_{model}$, then the task of discriminator is to distinguish between the input coming from $P_{data}$ ($x_{real}$) and generated by generator $x_{fake}$. This is achieved by maximizing the cost function $C_{D}(.;\theta_{G}, \theta_{D})$. Equation \eqref{eq:eq2} and \eqref{eq:eq3} describes $C_G$ and $C_D$. This arrangement sets up a min-max game between $G$ and $D$.
\begin{align}
\scriptstyle C_{G}(z; \mathbf{\Theta}) &= \scriptstyle -\mathrm{E_{z \sim P_{prior}}}\log{D(G(z))} \label{eq:eq2} \\
\scriptstyle C_{D}(x; \mathbf{\Theta}) &= \scriptstyle -\mathrm{E_{x \sim P_{data}}}\log{D(x)}
-\mathrm{E_{z \sim P_{prior}}}\log{(1-D(G(z)))}
\label{eq:eq3}
\end{align}
Here $\mathbf{\Theta} = \{ \theta_G, \theta_D \}$. Equation \eqref{eq:eq2} here uses the non-saturating heuristic in order to train network efficiently  \cite{gan_tut}.

\subsection{Batch effect correction using GANs}
Consider two data batches, source ($S$) and target ($T$) coming from the identical initial samples but prepared under different conditions. It is assumed that the experiments were performed in a careful manner, this can minimize confounding and leave batch effect as the prime reason for variations across batches.
\begin{figure*}
\minipage{0.2\textwidth}
  \includegraphics[width=\linewidth]{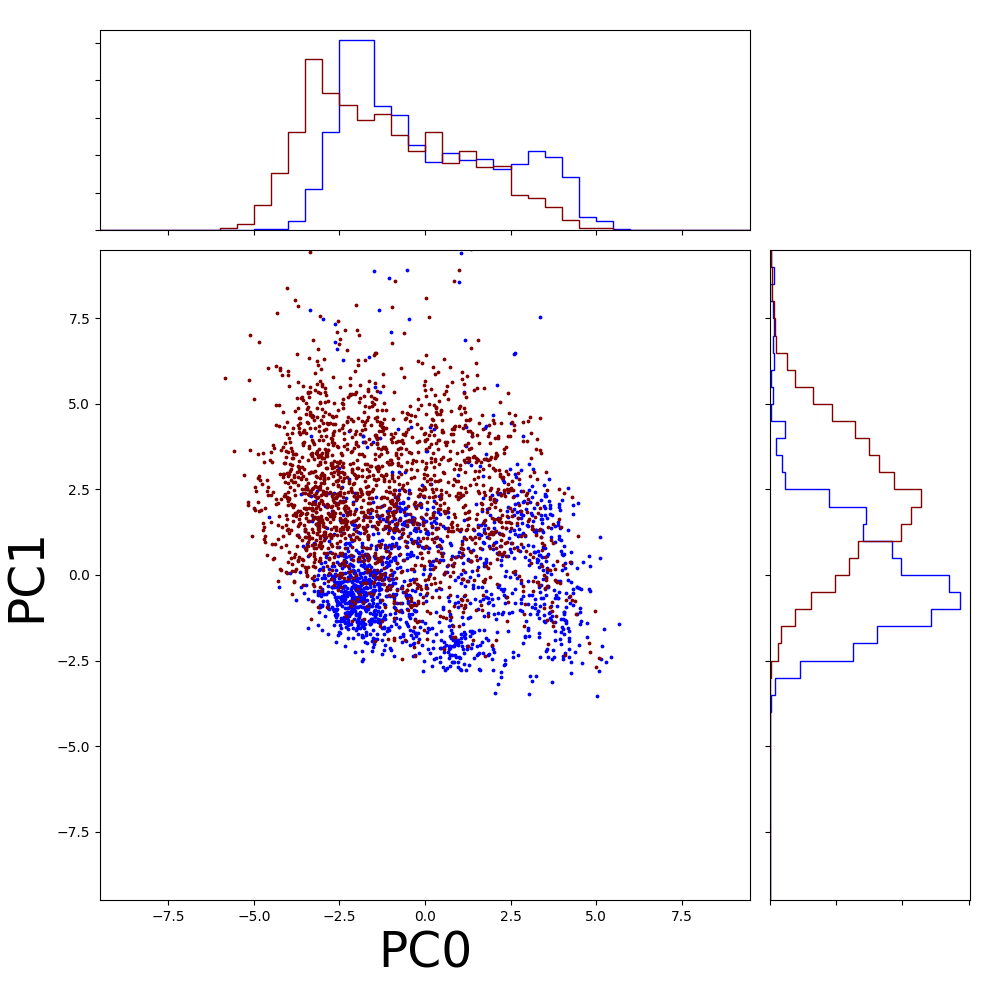}
  \centering{(a1)}
\endminipage\hfill
\minipage{0.2\textwidth}
  \includegraphics[width=\linewidth]{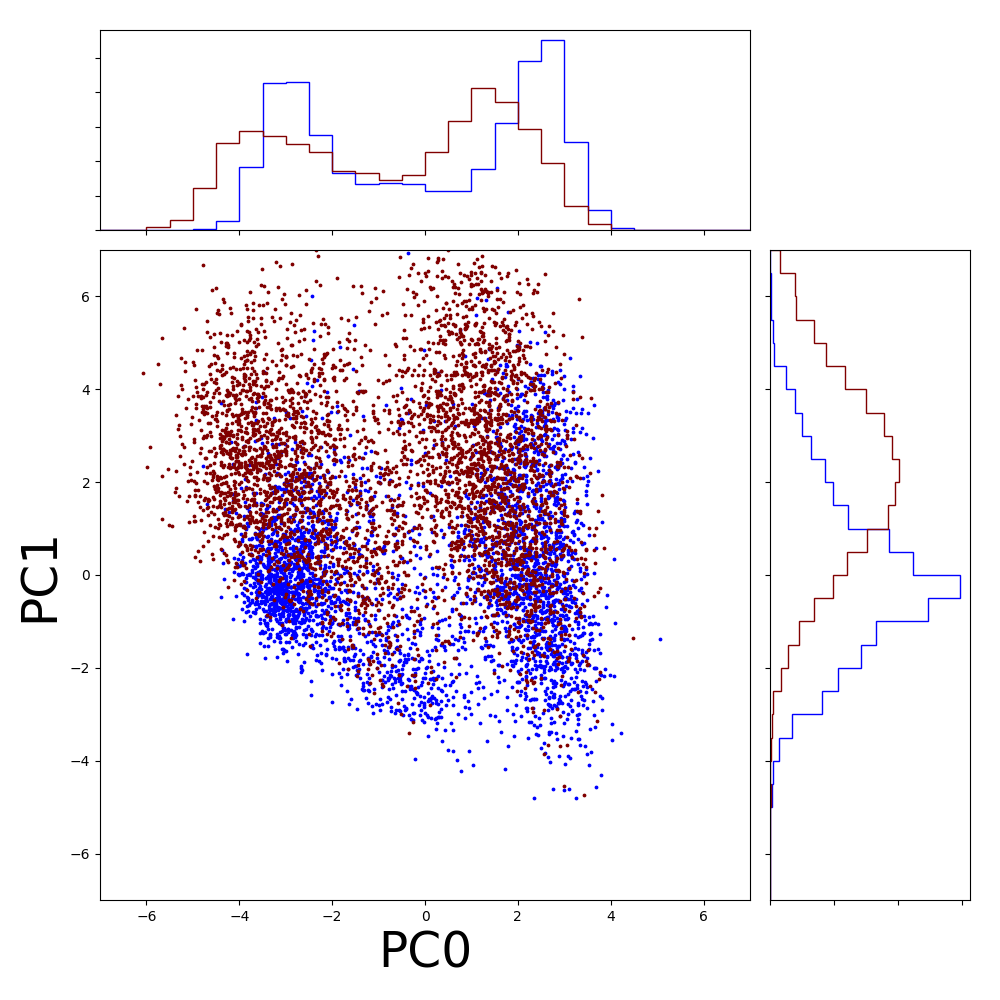}
  \centering{(b1)}
\endminipage\hfill
\minipage{0.2\textwidth}%
  \includegraphics[width=\linewidth]{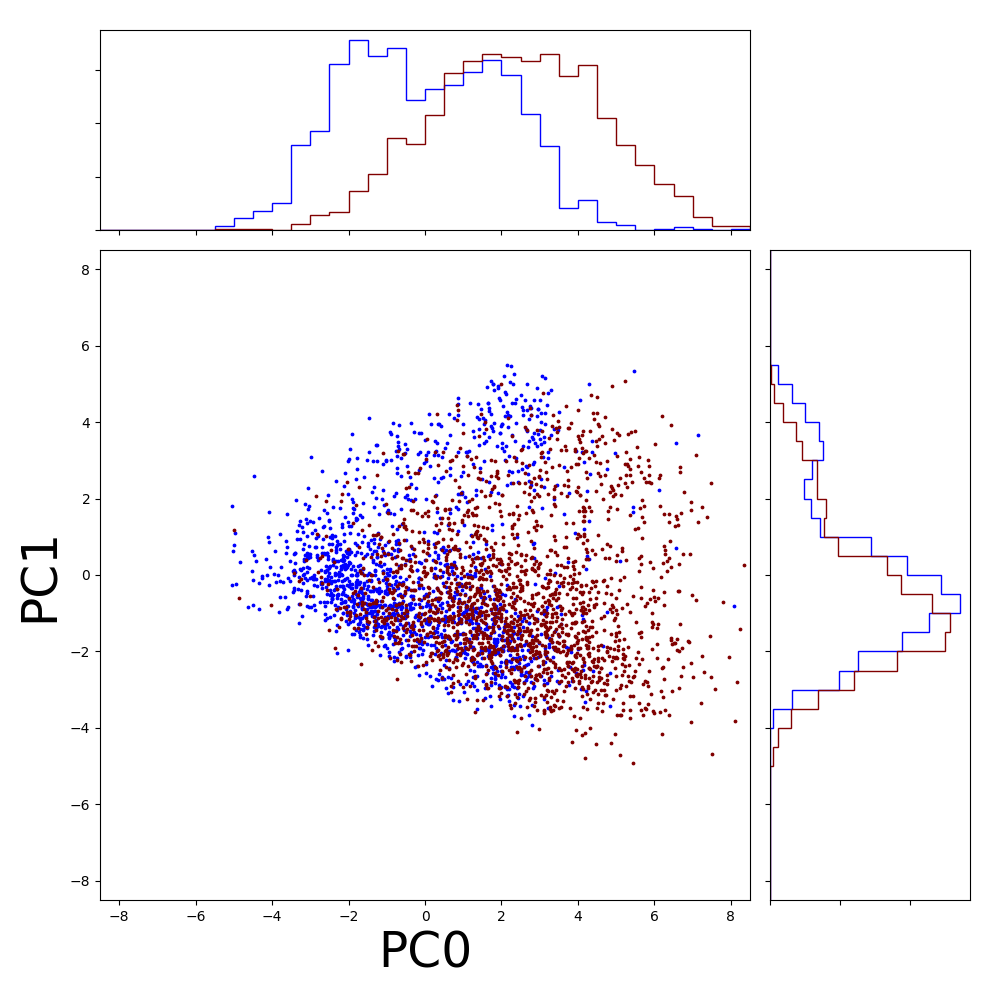}
  \centering{(c1)}
\endminipage\hfill
\minipage{0.2\textwidth}%
  \includegraphics[width=\linewidth]{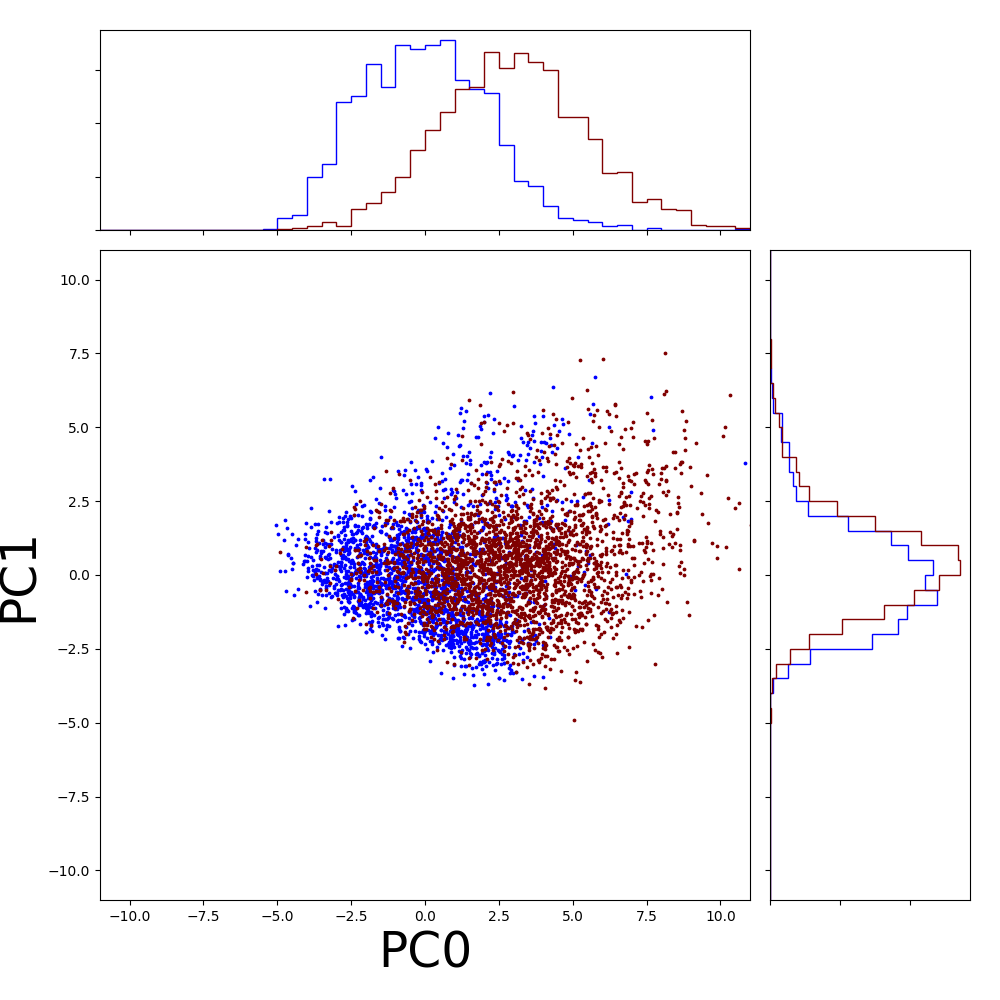}
  \centering{(d1)}
\endminipage

\minipage{0.2\textwidth}
  \includegraphics[width=\linewidth]{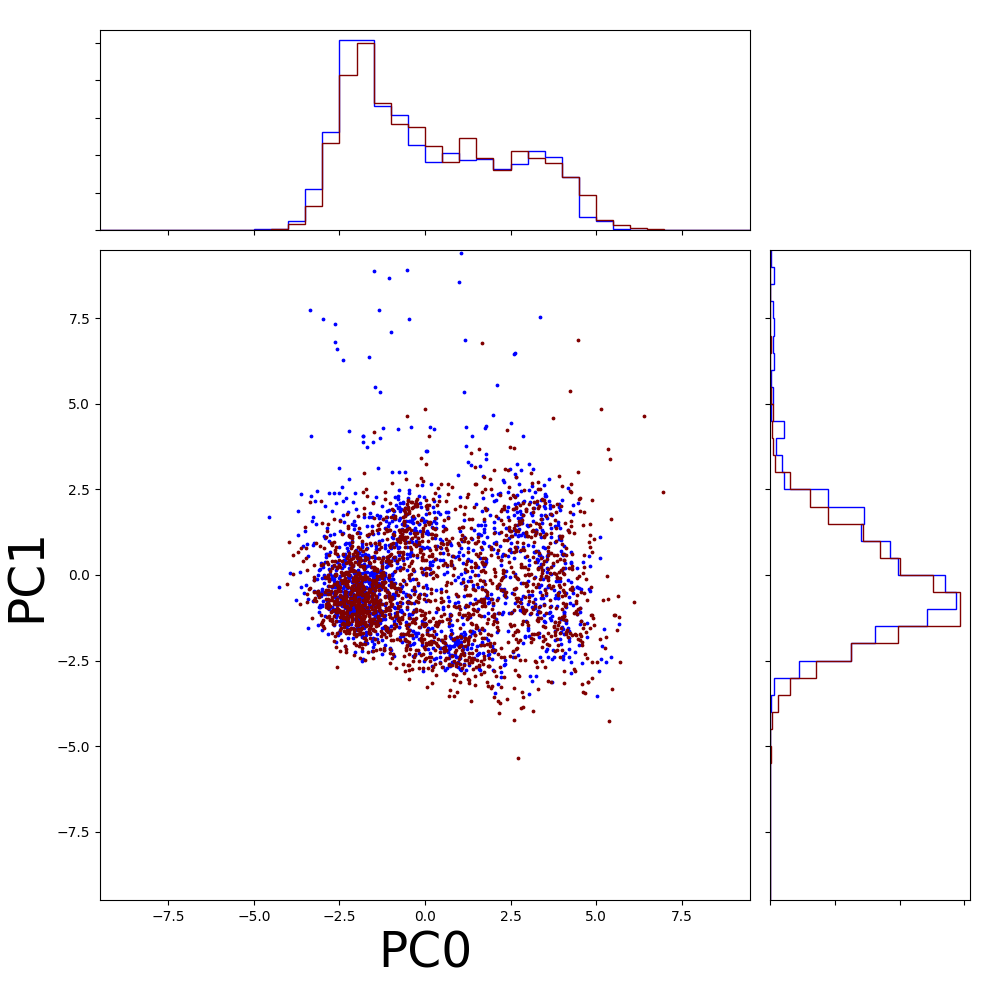}
  \centering{(a2)}
\endminipage\hfill
\minipage{0.2\textwidth}
  \includegraphics[width=\linewidth]{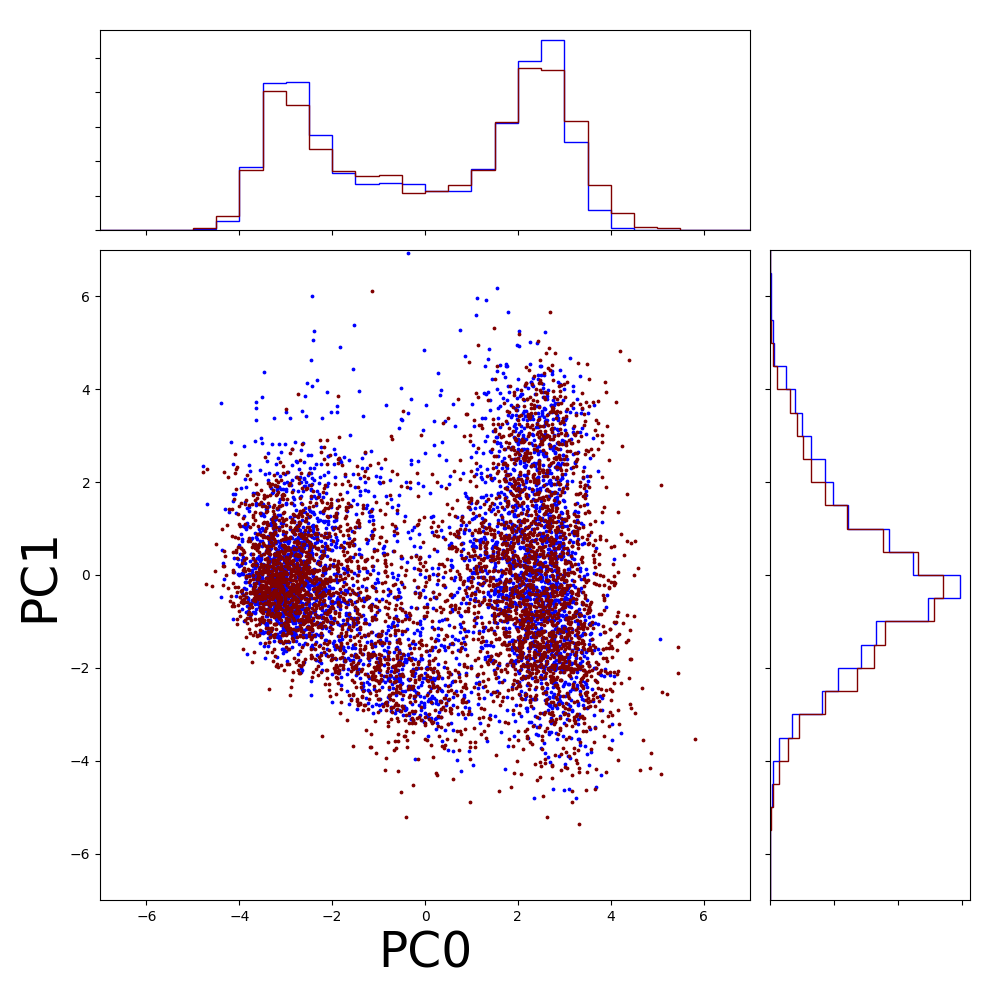}
  \centering{(b2)}
\endminipage\hfill
\minipage{0.2\textwidth}%
  \includegraphics[width=\linewidth]{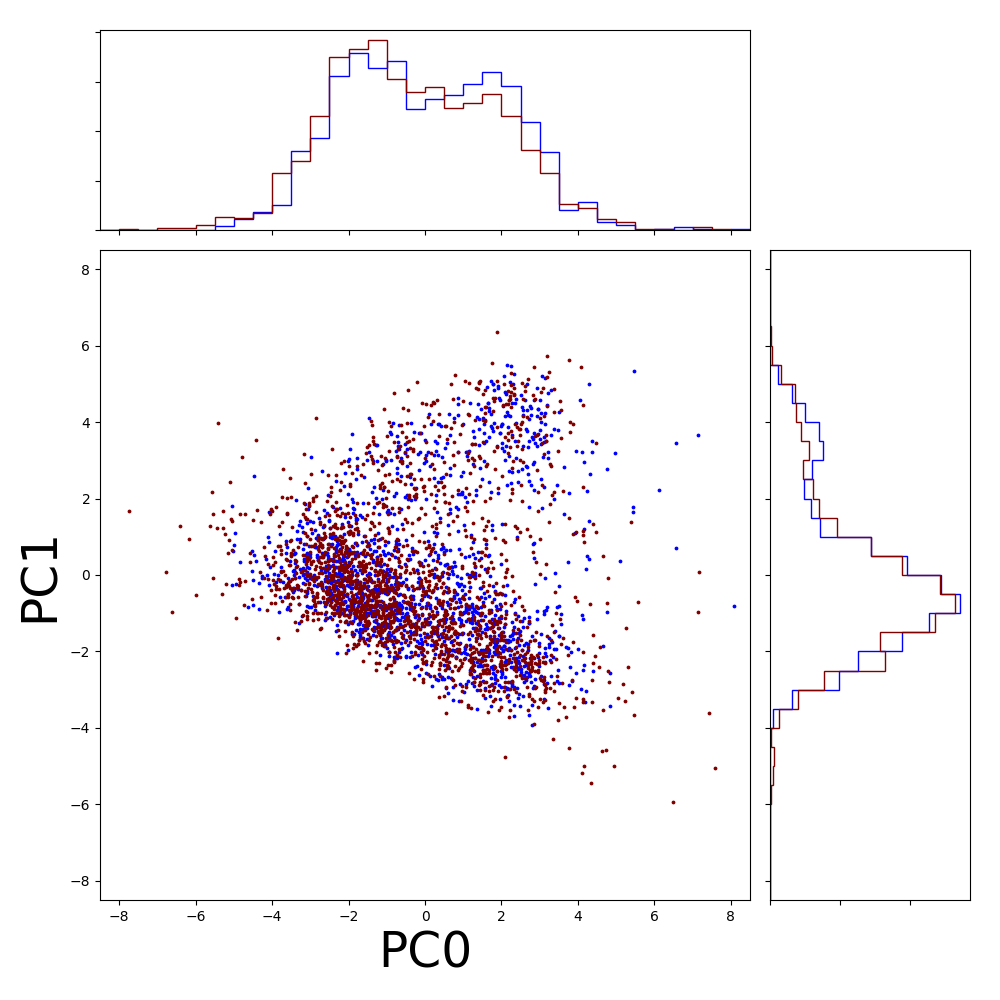}
  \centering{(c2)}
\endminipage\hfill
\minipage{0.2\textwidth}%
  \includegraphics[width=\linewidth]{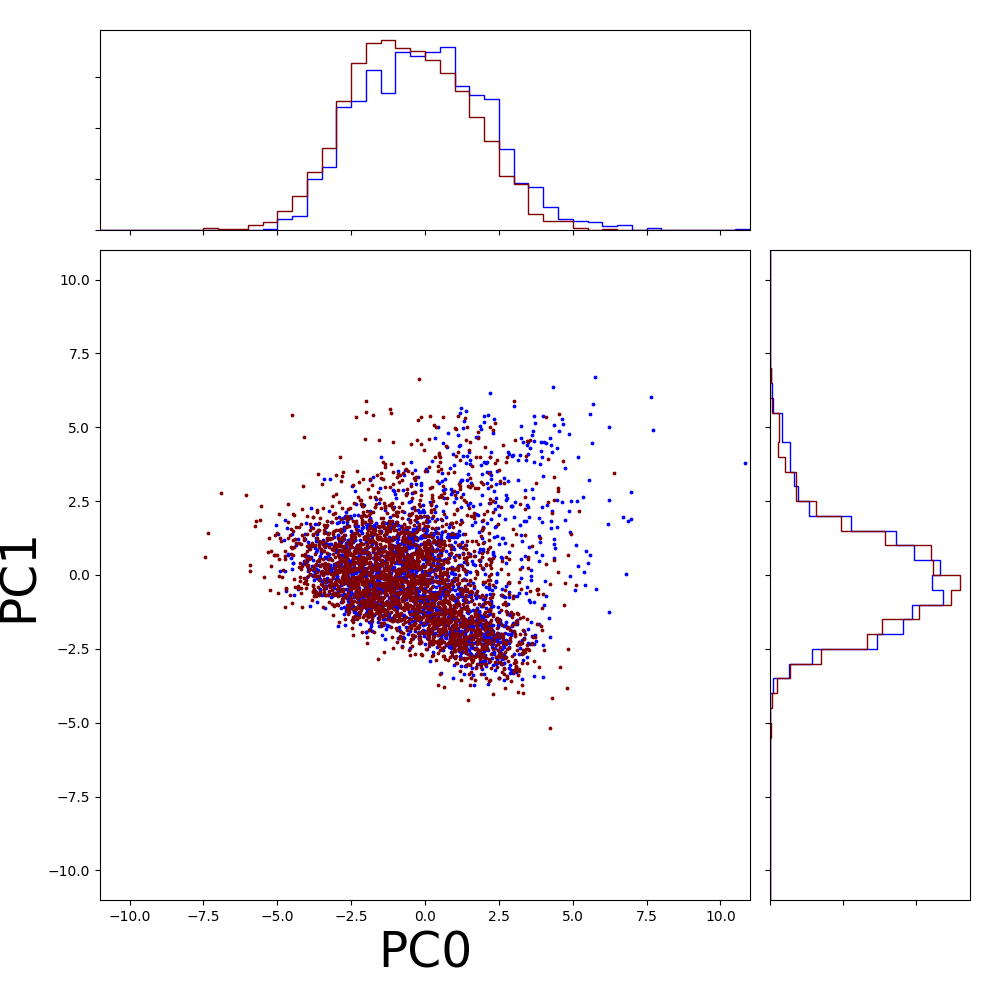}
  \centering{(d2)}
\endminipage

\caption{PCA plots of four source-target pairs. Blue and Maroon represents source and target respectively. Top-row [a1-d1] represents data before calibration, each plot has data from the same patient under same conditions but batches were created on two different days. Note the variations in batches. Bottom-row [a2-d2] represents data after calibration using GAN}
\label{fig:res}
\end{figure*}
Removal of these batch effects can be posed as problem of finding a map $\Psi$ such that
\begin{equation}
\Psi(x) \in T \;\;\; \forall x \in S   
\end{equation}
To use a GAN for finding such a mapping $\Psi$, the problem is framed in the following manner.

Let source ($S$) and target ($T$) have an underlying distribution $P_{S}$ and $P_{T}$ (both of which are unknown). GAN is set up such that the generator takes input $z \sim P_{S}$ and produces output $x_{fake} \sim P_{model}$. Discriminator learns to discriminate between $x_{real} \sim P_{T}$ and $x_{fake} \sim P_{model}$. Then $\Psi$ can be obtained by training the GAN and using the generator $G$ to produce $x_{fake}$ from $z$, networks are trained to get the parameters ($\theta_{G}^*$, $\theta_{D}^*$)
\begin{align}
\scriptstyle \theta_{D}^* &= \scriptstyle \underset{\theta_D}{argmin}[-(\mathrm{E_{x \sim P_{T}}}\log{D(x)}
+\mathrm{E_{z \sim P_{S}}}\log{(1-D(G(z)))})] \label{eq:eq4} \\
\scriptstyle \theta_{G}^* &= \scriptstyle \underset{\theta_G}{argmin}[-\mathrm{E_{z \sim P_{S}}}\log{D(G(z))}]
\label{eq:eq5}
\end{align}
The network is trained in such a manner that while updating the weights of discriminator, the weights of the generator are fixed and vice-versa as discussed in next section.

\section{Experiments and Results}
\begin{figure*}
\minipage{0.2\textwidth}
  \includegraphics[width=\linewidth]{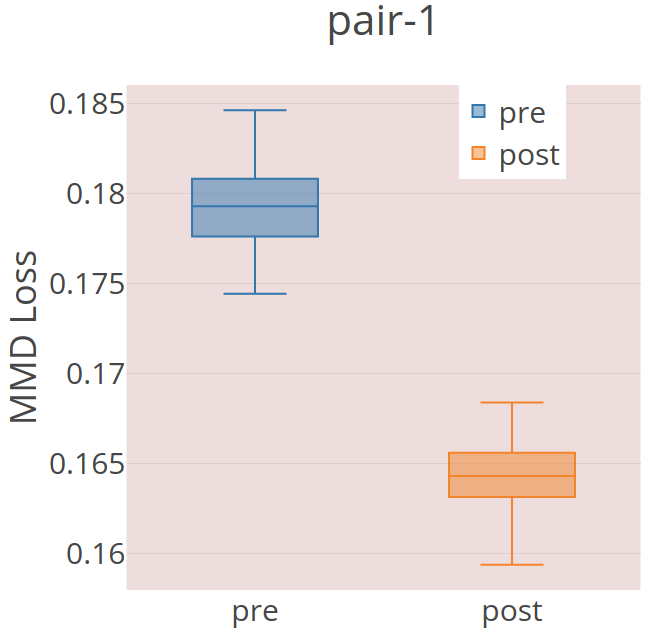}
  \centering{(a1)}
\endminipage\hfill
\minipage{0.2\textwidth}
  \includegraphics[width=\linewidth]{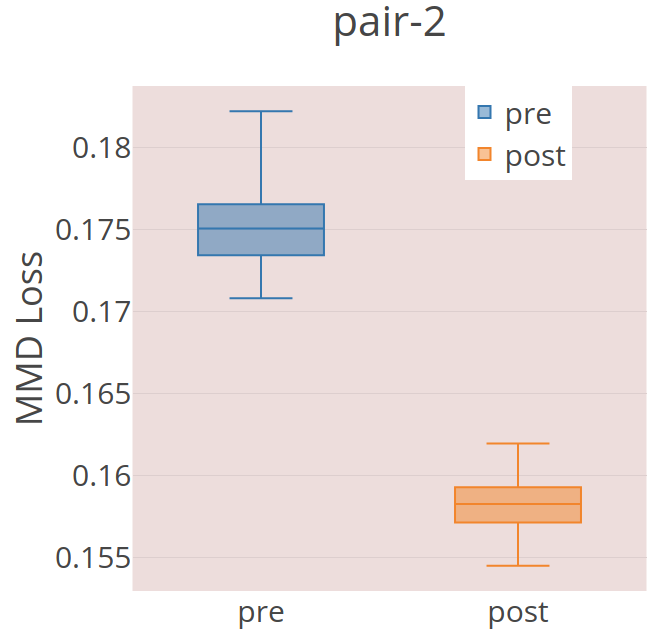}
  \centering{(b1)}
\endminipage\hfill
\minipage{0.2\textwidth}%
  \includegraphics[width=\linewidth]{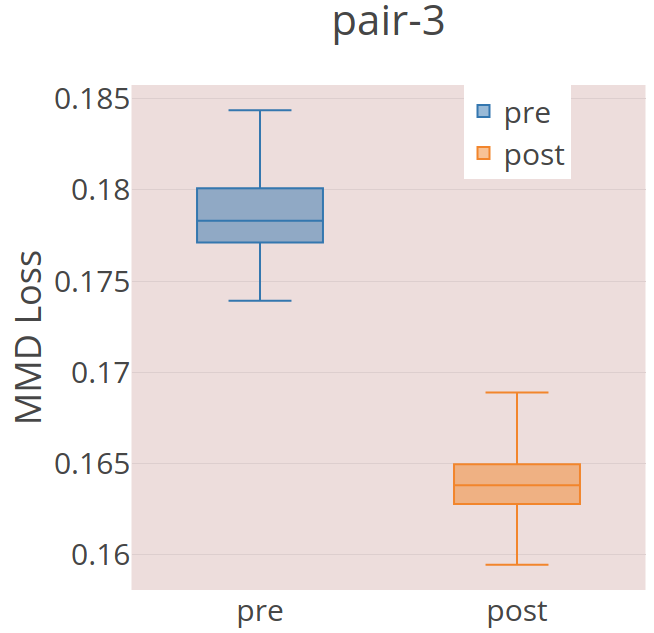}
  \centering{(c1)}
\endminipage\hfill
\minipage{0.2\textwidth}%
  \includegraphics[width=\linewidth]{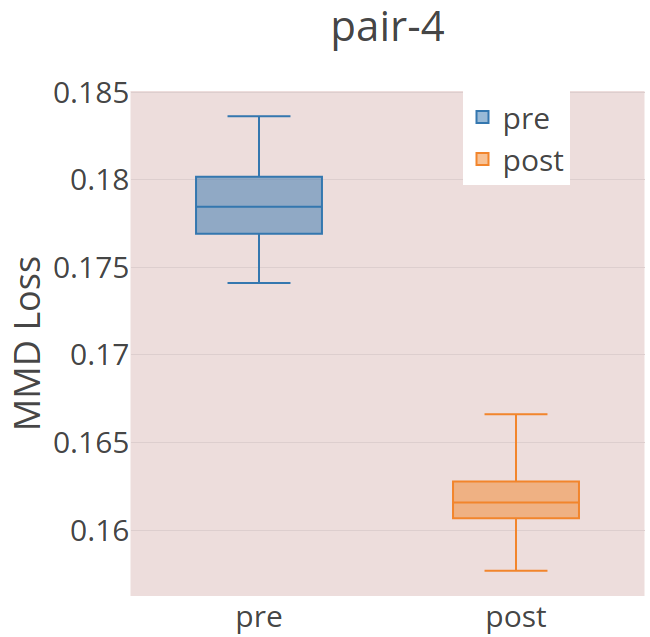}
  \centering{(d1)}
\endminipage

\caption{Box plots comparing the MMD Loss between the source and target pairs, pre-calibration and post-calibration using GAN. $256$ points were sampled $100$ times from source and targets to calculate and plot the loss}
\label{fig:res1}
\end{figure*}

\subsection{Dataset}
\textbf{Mass Cytometry} is a process to analyze the properties of the cells by labelling different proteins of the cell with antibodies conjugated with isotopically pure metals. These labelled cells are then nebulized and metal conjugated antibodies are ionised, such metal signals are analyzed to to determine the properties of the cells.

The dataset consists of Peripheral Blood Mononuclear Cells (PBMC) samples from two Multiple Sclerosis (MS) patients. Samples were collected on two different days, 90 days apart (baseline and after Gilenya treatment). Samples were cryopreserved. Each sample was divided in two batches on two different days and one of them was stimulated with PMA/ionomycin. Batches prepared on first and second day were treated as target and source respectively. Preprocessing of the dataset was done as explained in \cite{shaham2017removal,finck2013normalization}.

\subsection{Architecture and Adversarial Training}
\begin{figure}
    \centering
    \includegraphics[width=0.8\linewidth]{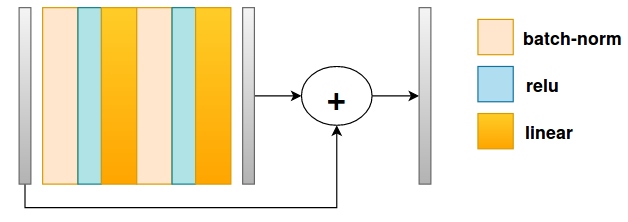}
    \caption{Generator with batch-norm, linear layers and residual skip connections}
    \label{fig:gen}
\end{figure}

\begin{figure}
    \centering
    \includegraphics[width=0.8\linewidth]{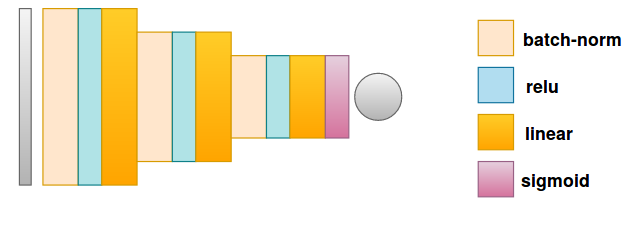}
    \caption{Discriminator with batch-norm and linear layers}
    \label{fig:dis}
\end{figure}

Proposed framework consists of two networks, a generator $G$ and a discriminator $D$. Generator consists of batch-norm layers \cite{bn}, linear layers and residual skip~\cite{he2016deep} connections as these will be essential to learn a mapping which is close to an identity mapping~\cite{shaham2017removal}. Generator is required to map a collection of points from target to source batches, both of which are 25-dimensional therefore $G$ is designed such that input-output dimensions match. $G$ is shown in Figure~\ref{fig:gen}. 
Discriminator $D$ is designed such that it takes the input form higher dimensional space (25-dimensional in this dataset) and outputs a scalar quantity. It consists of batch-norm layers and linear layers and applies sigmoid activation before outputting the scalar. $D$ is shown in Figure~\ref{fig:dis}.

To train both the networks, at every iteration $256$ points were sampled randomly from target and source batches. The group of points from source batch is passed through the generator which outputs ``fake samples'', this along with ``real samples'' (from target batch) is passed through the discriminator with labels of ``real samples'' and ``fake samples'' set to $1$ and $0$ respectively. Discriminator optimizes for equation \ref{eq:eq3}. Generator uses the ``fake samples'' with labels set to $1$ and optimizes for Equation \ref{eq:eq2}. This is done in sequence hence while training $D$ parameters of $G$ are fixed and vice-versa. Adam \cite{kingma2014adam} optimizer with (learning rate, $\beta_1$, $\beta_2$) set to ($1e-3,0.9, 0.999$) respectively.

\subsection{Results}
Figure~\ref{fig:res} shows the PCA plots of the dataset (using first two major principle components). In total there were eight batches (2 patients $\times$ 2 conditions $\times$ 2 days) and four source-target pairs corresponding to 2 patients and 2 conditions, source being day-1 and target day-2. To quantitatively measure the batch effect MMD Loss was used with Gaussian kernels~\cite{shaham2017removal}. Results of Figure~\ref{fig:res1} show that the batch effects present in the raw dataset was significantly removed after passing the data through trained generator.

\section{Conclusions}
A novel solution to correct for batch effects is proposed using Generative Adversarial Networks. Results conclude that GANs can be used for this task without performing any explicit kernel based computation and thus reducing the need for domain knowledge of an expert or the intuition on an analyst to define the kernel or other hyper-parameters.

\bibliographystyle{IEEEbib}
\bibliography{refs}

\begin{thebibliography}{10}

\bibitem{doi:10.1093/bioinformatics/btt480}
Sarah~E. Reese, Kellie~J. Archer, Terry~M. Therneau, Elizabeth~J. Atkinson,
  Celine~M. Vachon, Mariza de~Andrade, Jean-Pierre~A. Kocher, and Jeanette~E.
  Eckel-Passow,
\newblock ``A new statistic for identifying batch effects in high-throughput
  genomic data that uses guided principal component analysis,''
\newblock {\em Bioinformatics}, vol. 29, no. 22, pp. 2877--2883, 2013.

\bibitem{shaham2017removal}
Uri Shaham, Kelly~P Stanton, Jun Zhao, Huamin Li, Khadir Raddassi, Ruth
  Montgomery, and Yuval Kluger,
\newblock ``Removal of batch effects using distribution-matching residual
  networks,''
\newblock {\em Bioinformatics}, vol. 33, no. 16, pp. 2539--2546, 2017.

\bibitem{Dziugaite:2015:TGN:3020847.3020875}
Gintare~Karolina Dziugaite, Daniel~M. Roy, and Zoubin Ghahramani,
\newblock ``Training generative neural networks via maximum mean discrepancy
  optimization,''
\newblock in {\em Proceedings of the Thirty-First Conference on Uncertainty in
  Artificial Intelligence}, 2015, UAI'15, pp. 258--267.

\bibitem{li2015generative}
Yujia Li, Kevin Swersky, and Rich Zemel,
\newblock ``Generative moment matching networks,''
\newblock in {\em International Conference on Machine Learning (ICML)}, 2015,
  pp. 1718--1727.

\bibitem{gan_tut}
Ian Goodfellow,
\newblock ``Generative adversarial networks,''
\newblock in {\em Advances in Neural Information Processing Systems (NIPS)
  Tutorial}, 2016.

\bibitem{goodfellow2014generative}
Ian Goodfellow, Jean Pouget-Abadie, Mehdi Mirza, Bing Xu, David Warde-Farley,
  Sherjil Ozair, Aaron Courville, and Yoshua Bengio,
\newblock ``Generative adversarial nets,''
\newblock in {\em Advances in neural information processing systems (NIPS)},
  2014, pp. 2672--2680.

\bibitem{finck2013normalization}
Rachel Finck, Erin~F Simonds, Astraea Jager, Smita Krishnaswamy, Karen Sachs,
  Wendy Fantl, Dana Pe'er, Garry~P Nolan, and Sean~C Bendall,
\newblock ``Normalization of mass cytometry data with bead standards,''
\newblock {\em Cytometry Part A}, vol. 83, no. 5, pp. 483--494, 2013.

\bibitem{bn}
Christian~Szegedy Sergey~Ioffe,
\newblock ``Batch normalization: Accelerating deep network training by reducing
  internal covariate shift,''
\newblock in {\em ICML}, 2015.

\bibitem{he2016deep}
Kaiming He, Xiangyu Zhang, Shaoqing Ren, and Jian Sun,
\newblock ``Deep residual learning for image recognition,''
\newblock in {\em Proceedings of the IEEE conference on computer vision and
  pattern recognition}, 2016, pp. 770--778.

\bibitem{kingma2014adam}
Diederik~P Kingma and Jimmy Ba,
\newblock ``Adam: A method for stochastic optimization,''
\newblock in {\em International Conference on Learning Representations (ICLR)},
  2015.

\end{thebibliography}

\end{document}